\documentclass{article}
\usepackage{iclr2027_conference,times}
\iclrfinalcopy

\usepackage[utf8]{inputenc}
\usepackage[T1]{fontenc}
\usepackage{microtype}
\usepackage{amsmath,amssymb}
\usepackage{booktabs}
\usepackage{array}
\usepackage{graphicx}
\usepackage{longtable}
\usepackage{float}
\usepackage{xcolor}
\usepackage{hyperref}
\usepackage{url}
\hypersetup{hidelinks}
\usepackage{placeins}
\usepackage{tikz}
\usetikzlibrary{arrows.meta}
\title{ProteinJEPA: Latent prediction improves\\protein language model pretraining}

\author{
  Dan Ofer$^{1}$\thanks{Correspondence to \texttt{dan.ofer@mail.huji.ac.il}.} \quad
  Michal Linial$^{1}$ \quad
  Dafna Shahaf$^{2}$ \\[0.4em]
  $^{1}$Department of Biological Chemistry \quad
  $^{2}$School of Computer Science and Engineering \\
  The Hebrew University of Jerusalem
}
\begin{document}
\maketitle

\begin{abstract}
Protein language models are trained primarily with masked language modeling
(MLM), which predicts masked amino-acid identities. Joint-embedding predictive
architectures (JEPA) instead predict latent representations, but have not been
applied to proteins. ProteinJEPA supplements MLM with a cosine
loss for predicting the half-depth hidden states of a teacher given the unmasked sequence. On 19 tasks,
with ESM2 at 35M and 150M parameters and three pretraining seeds, MLM+JEPA
outperforms compute-matched and step-matched MLM-only continued training in 78 and 76 of 114 comparisons (14 losses, 22
ties). 
The median compute-matched gain is
$+0.0106$ on structure- and homology-sensitive tasks versus $+0.0041$
elsewhere, led by SCOPe-40 retrieval and remote homology with improvements of 6.1 percentage points in Recall@1 and 2.7 points in accuracy, respectively. Gains on these tasks increase with model size from 8M to 150M. Against the off-the-shelf checkpoint,
MLM+JEPA wins 81 of 114 comparisons (median $+0.0068$) without improving MLM loss.
 In random initialization the gain is smaller and replicates inconsistently across seeds ($p{=}0.059$). The same recipe improves the
causal ProGen3 model, beating a compute-matched next-token-prediction control
on 12 of 16 tasks. Ablations show that cosine loss beats mean squared error, while adding
shallower targets removes most of the task gain. JEPA-only
training collapses downstream performance: latent prediction complements MLM rather than replacing it.
Code: \url{https://anonymous.4open.science/r/protJepa-FF24}.
\end{abstract}

\section{Introduction}

Masked language modeling (MLM) is the standard self-supervised objective for
protein language models
\citep{brandes_proteinbert_2022,lin_evolutionary-scale_2023,rives_biological_2019,vieira_medium-sized_2025}:
mask residues and predict their identities. Joint-embedding predictive
architectures (JEPA)
\citep{lecun_path_2022,balestriero_lejepa_2025,huang_llm-jepa_2025,schmidhuber_discovering_1993}
provide a different signal: predicting a hidden representation of the unmasked
input rather than the input tokens. Latent prediction has worked well in
vision, video, language, and DNA, but has not been applied to proteins \citep{huang_llm-jepa_2025,balestriero_lejepa_2025,larey_jepa-dna_2026}.

We ask: \emph{can latent prediction improve protein language model
pretraining?} To our knowledge, ProteinJEPA is the first application of
JEPA-style latent prediction to protein language models. We find it complements
rather than replaces MLM: the model jointly predicts masked amino-acid identities and the half-depth hidden states of an exponential-moving-average (EMA) teacher (Sec.~\ref{sec:method}). Our controlled study
applies the recipe to ESM2 at three model sizes (8M, 35M, and 150M; three
independent pretraining seeds at 35M and 150M), continued from pretrained
checkpoints and trained from random initialization, against step-matched and
compute-matched MLM-only controls and the off-the-shelf checkpoint. We extend the recipe to the causal ProGen3-112m model, trained by next-token prediction (NTP). A second, wall-clock-matched study over five
backbones tests the breadth of the finding (Sec.~\ref{sec:breadth}).

\begin{figure}[H]
\centering
\includegraphics{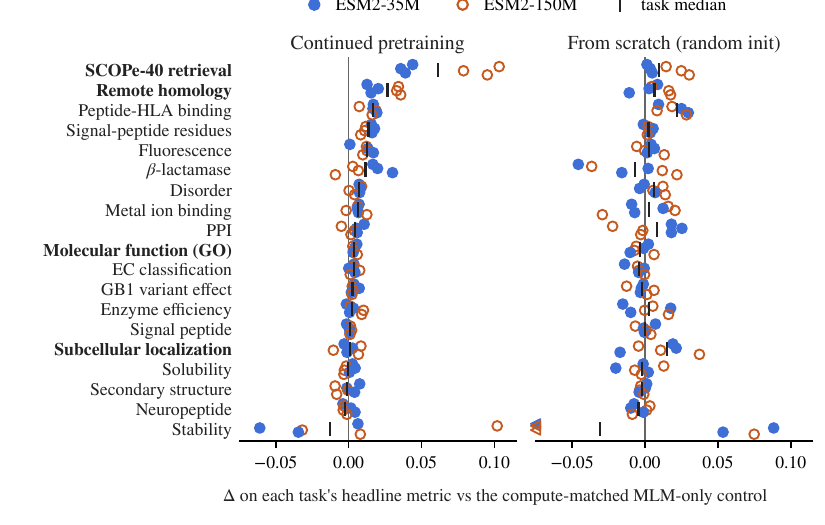}
\caption{Controlled study: MLM+JEPA minus compute-matched MLM-only on each
task's headline metric; positive values favor MLM+JEPA. Each row shows three
pretraining seeds at each scale (35M filled, 150M hollow) and their median
(black tick), ordered by the continued-pretraining median. Bold labels mark the
structure/homology subset (Sec.~\ref{sec:setup}). Triangles 
mark three off-scale from-scratch stability differences (
$-0.11$ to $-0.25$). Metrics are defined in Table~\ref{tab:task-sources}.}
\label{fig:v3-pertask}
\end{figure}

\paragraph{Contributions.}
\begin{itemize}
  \setlength\itemsep{0.1em}
  \item \textbf{Latent prediction improves continued pretraining under matched
  budgets} (Table~\ref{tab:v3-results}). MLM+JEPA beats step- and
  compute-matched MLM-only in every pretraining seed, and is 81/20/13 against
  the off-the-shelf checkpoint where MLM-only continuation is 52/35/27. The
  same holds for the causal ProGen3 model (Sec.~\ref{sec:ntp}).
  \item \textbf{The gain concentrates on structural tasks and grows with model size.}
  SCOPe-40 retrieval and remote homology gain most, and their gains grow from
  8M to 150M (Appendix~\ref{app:scale-trend}); the structure/homology subset's
  median gain over the compute-matched control is $+0.0106$, versus $+0.0041$
  on the other fifteen tasks (Fig.~\ref{fig:v3-pertask},
  Sec.~\ref{sec:structure}). Peptide--HLA binding, signal-peptide residues,
  fluorescence, and $\beta$-lactamase also gain in at least five of six runs,
  and per-residue probes improve.
  \item \textbf{Latent prediction complements, but does not replace MLM/NTP.}
  JEPA-only continuation degrades every pretrained backbone, losing 17 of 19
  tasks on ESM2-35M (median $-0.124$; Sec.~\ref{sec:objective-ablations}). It also underperforms MLM from random initialization.
  \item \textbf{Continued pretraining is where the gain is largest.} MLM+JEPA
  wins 78 and loses 14 of 114 comparisons when continuing pretraining, but
  wins 54 and loses 38 from random initialization (Sec.~\ref{sec:scratch}).
  \item \textbf{Recipe ablations.} Cosine loss beats mean squared error.
  Masked- and all-position targets have similar downstream performance, and adding shallower
  targets to the half-depth one removes most of the structure-task gain
  (Secs.~\ref{sec:ladder} and~\ref{sec:objective-ablations}).
\end{itemize}

\section{Related Work}

\paragraph{Protein language models.}
Protein language models rely predominantly on token prediction, masked in
ESM2~\citep{lin_evolutionary-scale_2023},
ProtTrans~\citep{elnaggar_prottrans_2020}, and
AMPLIFY~\citep{fournier_protein_2026} and causal in
ProGen2~\citep{nijkamp_progen2_2022}.
ProteinBERT~\citep{brandes_proteinbert_2022,michael-pitschaze_detecting_2024},
SaProt~\citep{su_saprot_2024}, and ESM-3~\citep{hayes_simulating_2025} add
function or structure channels. Our question is different: \textit{what does a
self-supervised latent target add to protein language model pretraining, and
is it worth its compute overhead?}

\paragraph{Joint-embedding predictive learning.}
I-JEPA~\citep{assran_self-supervised_2023} and
V-JEPA~\citep{bardes_revisiting_2024} established latent prediction in vision
and video; LLM-JEPA~\citep{huang_llm-jepa_2025} and
JEPA-DNA~\citep{larey_jepa-dna_2026} extend the idea to language and DNA. A
latent loss can be satisfied by mapping every input to the same vector
(collapse); JEPA methods prevent this with a slowly updated target encoder or a
regularizer that keeps embeddings spread out, such as
VICReg~\citep{bardes_vicreg_2022}, SIGReg~\citep{balestriero_lejepa_2025},
and VISReg~\citep{wu_visreg_2026}.

\paragraph{Latent targets for proteins.}
Continued pretraining further trains an existing model with its original
objective~\citep{gururangan_dont_2020} or with new task-specific
objectives~\citep{schmirler_fine-tuning_2024}. Targets other than masked tokens
used to pretrain protein models include functional annotations, distillation,
and contrastive
fine-tuning~\citep{geffen_distilprotbert_2022,ofer_protsent_2026,brandes_proteinbert_2022,heinzinger_contrastive_2022};
we instead add a data2vec-style EMA-teacher target~\citep{baevski_data2vec_2022}
to MLM/NTP and, in the wall-clock study, a teacher-free variant regularized with
SIGReg inspired by LeJEPA~\citep{balestriero_lejepa_2025} (Sec.~\ref{sec:method}).

\begin{samepage}
\section{Method}
\label{sec:method}

ProteinJEPA adds a latent prediction loss to MLM at the masked positions:
\begin{equation}
  \mathcal{L} = \mathcal{L}_{\text{MLM}}
              + \lambda\,\mathcal{L}_{\text{JEPA}}^{\text{masked}}
              + \alpha\,\mathcal{L}_{\text{reg}},
  \label{eq:loss}
\end{equation}
where $\mathcal{L}_{\text{MLM}}$ is masked-token cross-entropy and
$\mathcal{L}_{\text{JEPA}}^{\text{masked}}$ averages one minus cosine similarity
between student predictions and targets from the \emph{unmasked} input.
Both losses use the same masked positions. The student (the model being
trained) passes its final-layer hidden states through a small feed-forward
predictor with SwiGLU gating; predictions and targets are layer-normalized
before comparison.
$\mathcal{L}_{\text{reg}}$ discourages collapse, and $\lambda$ and $\alpha$
weight the latent and regularizer losses (Fig.~\ref{fig:method}).
\end{samepage}

\begin{figure}[htbp]
\centering
\resizebox{\linewidth}{!}{
\begin{tikzpicture}[
  font=\small,
  box/.style={draw, rounded corners=2pt, align=center, inner sep=4pt, minimum height=2.4em},
  loss/.style={box, fill=black!8},
  arr/.style={-{Latex[length=1.5mm]}, semithick},
  lbl/.style={font=\footnotesize, inner sep=1.5pt, fill=white},
  x=1cm, y=1cm,
]
\node[box] (mask) at (0,1.2) {sequence $x$\\15\% masked};
\node[box] (clean) at (0,-2.4) {sequence $x$\\unmasked};
\node[box, minimum width=2.6cm] (student) at (3.1,1.2) {student encoder\\$L$ layers};
\node[box] (mlm) at (6.6,1.2) {MLM head};
\node[loss] (ce) at (10.2,1.2) {cross-entropy\\at masked positions};
\node[box] (pred) at (6.6,-0.4) {SwiGLU predictor\\$+$ LayerNorm};
\node[loss] (cos) at (10.2,-0.4) {cosine loss\\at masked positions};
\node[loss] (reg) at (6.6,-2.4) {VISReg\\maintains variation};
\node[box, minimum width=2.6cm] (teacher) at (3.1,-2.4)
  {EMA teacher\\layers $1$ to $L/2$\\no gradient};

\draw[arr] (mask) -- (student);
\draw[arr] (student) -- (mlm);
\draw[arr] (mlm) -- (ce);
\draw[arr] ([yshift=-0.25cm]student.east) -- (4.9,0.95) |- (pred.west);
\node[lbl, anchor=south west] at (4.9,0.2) {layer $L$};
\draw[arr] (pred) -- (cos);
\draw[arr] (pred) -- (reg);
\draw[arr] (clean) -- (teacher);
\draw[arr] (teacher.south) -- (3.1,-3.65) -- (10.2,-3.65) -- (cos.south);
\node[lbl, anchor=south] at (7.9,-3.65) {target $+$ LayerNorm};
\draw[arr, dashed] (student.south) -- (teacher.north)
  node[lbl, midway, left, align=right] {EMA\\weight update};
\end{tikzpicture}}
\caption{Controlled ProteinJEPA continuation. The student learns from token
prediction and a half-depth latent target from an EMA teacher. Both prediction
losses use masked positions. VISReg keeps predicted embeddings diverse;
the dashed arrow updates teacher weights, without backpropagation.}
\label{fig:method}
\end{figure}

\paragraph{Why a half-depth target?}
MLM trains the final layer to recover residue identity. Intermediate layers
of protein language models can carry more structure and homology information
than the final
layer~\citep{ofer_language_2021,michael-pitschaze_detecting_2024,siji_words_2026,valeriani_geometry_2023,vig_bertology_2020}.
We therefore predict the half-depth hidden state of the teacher model (layer 6
of 12 at 35M, 15 of 30 at 150M).
The teacher is updated by averaging its weights with the student's after each
optimizer step and receives the original (unmasked) full input sequence;
gradients do not pass through the teacher. Truncating the target layer also
reduces its overhead: 17\% more compute than MLM-only, compared with about
37\% for a last-layer target.

\paragraph{Regularization.}
The controlled study combines the EMA teacher~\citep{baevski_data2vec_2022}
with VISReg (variance--invariance--sketching regularization;~\citealp{wu_visreg_2026}).
As a latent loss alone can be minimized by predicting the same vector for every
input (collapse), VISReg prevents this by projecting the predicted embeddings
onto random directions and penalizing each projection's sliced-Wasserstein
distance from a standard Gaussian, so predictions must stay spread
out; SIGReg (sketched isotropic Gaussian
regularization;~\citealp{balestriero_lejepa_2025,maes_leworldmodel_2026})
applies the same regularization with the Epps--Pulley statistic.

The wall-clock study instead uses the student's own gradient-blocked last layer
as the target, with SIGReg and per-backbone hyperparameters listed in
Table~\ref{tab:wallclock-hparams}.

\paragraph{Ablated objectives.}
The wall-clock study also includes \textbf{all-position MLM+JEPA},
which applies the latent loss to all non-padding positions, and
\textbf{JEPA-only}, which removes MLM entirely. Within the controlled study,
the \textbf{ladder} variant uses targets at layers 3--6 of 12 (35M) or 7, 9,
11, 13, and 15 of 30 (150M): the half-depth target plus shallower ones.
Teacher, regularizer, loss weight, and schedule are otherwise unchanged.

\section{Experimental Setup}
\label{sec:experiments}

We use two complementary studies. The \textbf{controlled study}, ESM2 against
step- and compute-matched MLM-only controls, is the primary evidence; the
\textbf{wall-clock study} (Sec.~\ref{sec:breadth}) covers five backbones at a
fixed 8\,h budget. Table~\ref{tab:two-studies} lists their differences, and
Sec.~\ref{sec:ntp} applies the controlled recipe to a causal model. Throughout,
an \emph{arm} is one training condition, such as MLM-only or MLM+JEPA.

\begin{table}[htbp]
\centering
\small
\caption{The two studies. Both headline recipes use MLM plus a cosine latent
loss at masked positions; they differ in matching, collapse control, and target
layer. $\lambda$ and $\alpha$ weight the latent and regularizer terms of
Eq.~\ref{eq:loss}.}
\label{tab:two-studies}
\begin{tabular}{@{}lll@{}}
\toprule
 & Controlled study & Wall-clock study (Sec.~\ref{sec:breadth}) \\
\midrule
Budget matched on   & steps, and separately compute & wall-clock, 8\,h \\
Backbones           & ESM2-35M, ESM2-150M           & five backbones \\
Pretraining seeds   & three, pooled                 & pooled, 5--8 runs per arm \\
Corpus              & UniRef50                      & UniRef30 \\
\midrule
Target              & EMA teacher, half depth       & last layer, gradients blocked \\
EMA decay           & $0.996 \rightarrow 0.9995$    & --- (no teacher network) \\
Collapse control    & VISReg, $\alpha{=}0.008$      & SIGReg, 256 projections \\
Predictor / $\lambda$ & 1-layer SwiGLU, $1.0$       & 2-layer SwiGLU, $0.45$ \\
Masking rate        & 15\,\% cont., 20\,\% scratch  & 25\,\% ESM2, 20\,\% other \\
\bottomrule
\end{tabular}
\end{table}

\subsection{Backbones}

The controlled study uses ESM2 at 35M and 150M parameters
\citep{lin_evolutionary-scale_2023}, each either continued from its
off-the-shelf checkpoint or trained from random initialization. An ESM2-8M
run with the same recipe (one pretraining seed) adds a third size
(Appendix~\ref{app:scale-trend}). We adapt the recipe
to ProGen3-112m \citep{bhatnagar_scaling_2025}, a causal next-token-prediction model
(Sec.~\ref{sec:ntp}). The wall-clock study uses ESM2-35M, ESM2-150M, AMPLIFY-120M
\citep{fournier_protein_2026} and ProteinBERT2-35M, a ProteinBERT-style encoder with a ModernBERT backbone
\citep{brandes_proteinbert_2022,warner_smarter_2024}.

\subsection{Training and evaluation setup}
\label{sec:setup}

\paragraph{Training.}
Sequences are truncated or padded to 512 tokens. Both arms of every MLM contrast
share the masking rate: 15\% for continued runs and 20\% from scratch in the
controlled study, and the per-backbone rates in
Table~\ref{tab:wallclock-hparams} for the wall-clock study. Runs use AdamW with weight decay 0.01, bfloat16 mixed precision, and FlashAttention-2
\citep{dao_flashattention-2_2023}; hardware details are in
Appendix~\ref{app:repro}.

The controlled ESM2 study trains on UniRef50~\citep{Suzek2015}, ESM2's original
training corpus, for 67{,}000 steps with an effective batch size of 1024---one
pass over the corpus. Peak learning rates are $2\times10^{-4}$ for continuation
and $1\times10^{-3}$ for training from scratch~\citep{geiping_cramming_2022},
reached after 2{,}000 warmup steps and decayed by a cosine schedule to 1\% of
peak.
The latent branch switches on after an MLM-only warm start: at step 2{,}000
with a 1{,}000-step ramp on $\lambda$ for continuation, and at step 20{,}000
with an 8{,}000-step ramp from scratch, allowing the model to learn from MLM
before introducing latent targets.

The wall-clock study trains on UniRef30~\citep{mirdita_uniclust_2017} with a
constant schedule after 1{,}000 warmup steps; the headline grid uses checkpoints
at $\{1,4,8\}$ hours. Per-checkpoint optimizer-step, sample, and token counts
for every run are in Appendix~\ref{app:training-coverage}.

\paragraph{Step and compute matching.}
The latent branch requires an additional target-encoder forward pass. Under
\emph{step matching}, both arms take 67{,}000 optimizer steps; the half-depth
MLM+JEPA recipe then uses 17\% more compute, measured as wall-clock time per
step on the same hardware. Under \emph{compute matching}, MLM-only receives
11{,}390 additional steps at its final learning rate (1\% of peak;
$2\times10^{-6}$ for continuation) to make up that 17\%.

Both arms are retrained at each seed, so each seed is an independent
replication. Within each pair, optimizer, schedule, learning rate, batch,
masking rate, and data seed are matched; the compute-matched control differs
only by its extra steps.

\paragraph{Benchmark tasks.}
The tasks fall into five groups. \emph{Fold and function}: SCOPe-40 retrieval \citep{Fox2014}
(retrieve a protein of the same structural family, defined in the Structural
Classification of Proteins, SCOPe; family fixes fold and superfamily), remote homology (fold class), molecular function (Gene Ontology,
GO, terms), EC classification (Enzyme Commission numbers) and subcellular
localization. \emph{Binding}: peptide--HLA binding, metal-ion binding, and
protein--protein interaction (PPI). \emph{Activity and variant effects}: enzyme
catalytic efficiency, $\beta$-lactamase activity, GB1
variant effect (binding fitness), fluorescence, and stability. \emph{Sequence properties}:
solubility, signal peptide, and neuropeptide precursor \citep{Karsenty2014}. \emph{Residue level}:
eight-state secondary structure, disorder, and signal-peptide residues. Tasks
are drawn from TAPE~\citep{rao_evaluating_2019}, \citep{xu_peer_2022}, FLIP \citep{dallago_flip_2021},
ProteinBERT~\citep{brandes_proteinbert_2022}, with further details in
Appendix~\ref{app:task-descriptions}, including metrics, and
per-task sources.

\paragraph{Evaluation.}
The controlled analysis uses 19 tasks: 18 linear probes (logistic or ridge
regression on frozen final-layer embeddings of the trained encoder),
mean-pooled for the 15 protein-level tasks and per residue for the three
residue-level ones, plus zero-shot SCOPe-40 retrieval
\citep{hubbard_scop_1999,Fox2014}. Recall@1 is the fraction of queries whose
nearest retrieved protein is in the same SCOPe family, using cosine similarity of
L2-normalized mean-pooled embeddings. The
wall-clock study uses a separate 19-task suite; Table~\ref{tab:task-sources}
lists both. To assess the half-depth-target hypothesis, we separately summarize
SCOPe-40 retrieval, remote homology, molecular function,
and subcellular localization as a structure/homology-sensitive subset.

\paragraph{Statistical reporting.}
Each task contributes one fixed headline metric, with higher scores indicating
better performance. We define $\Delta$ as the first-named arm minus its
comparison arm and summarize a contrast
by wins, losses, and ties (W/L/T), using $|\Delta|<0.002$ as the tie band, and
report one-sided exact sign tests (alternative $\Delta>0$) together with
one-sided Wilcoxon signed-rank tests for the controlled study. Table~\ref{tab:v3-results} pools the two model scales and three pretraining
seeds ($n{=}114$). Because these 114 comparisons reuse the same tasks, they are not independent;
per-seed, per-scale, and task-averaged tests support the continued-pretraining
result (Appendix~\ref{app:controlled-robustness}).

For the wall-clock study,
task-by-backbone comparisons use two-sided t-tests over independent
pretraining runs, with Benjamini--Hochberg correction across all 95; arms that share pretraining seeds
are paired (Appendix~\ref{app:repro}).

\section{Results}

\subsection{MLM+JEPA improves continued training under step and compute matching}
\label{sec:results-continued}

After one epoch of continued pretraining, MLM+JEPA beats step-matched MLM-only
on 76 of 114 task-by-run comparisons, with 15 losses and 23 ties ($p{<}0.0001$),
and the compute-matched control on 78, with 14 losses and 22 ties ($p{<}0.0001$;
Table~\ref{tab:v3-results}). The extra MLM updates do not close the gap. The
result replicates in every pretraining seed ($p{\le}0.0004$ per seed,
38 comparisons each; Appendix~\ref{app:controlled-robustness}).

The latent term does not improve MLM loss: training and validation MLM losses
are about 0.002 nats higher with MLM+JEPA at both scales and 0.005 nats higher
at 8M, and the extra compute-matching steps change validation loss by at most
0.005 nats (Appendix~\ref{app:controlled-robustness},
Figs.~\ref{fig:loss-curves} and~\ref{fig:val-loss};
Table~\ref{tab:8m-eval-loss}).

\begin{table}[!htbp]
\centering
\footnotesize
\setlength{\tabcolsep}{4pt}
\caption{Controlled study: ESM2-35M and 150M, three pretraining seeds, pooled
(19 tasks $\times$ 2 scales $\times$ 3 seeds $=$ 114 comparisons per row; the ladder
row is one seed and omits EC, 18 tasks $\times$ 2 scales $=$ 36). $\Delta$ is the first-named
arm minus the second and the median is over the row's comparisons; W/L/T counts
use $|\Delta|<0.002$ as the tie band; $p$-values are one-sided (alternative
$\Delta>0$). Step-matched MLM-only takes the same 67{,}000 steps as MLM+JEPA;
compute-matched MLM-only receives 11{,}390 extra MLM steps, matching MLM+JEPA's
measured 17\% overhead.}
\label{tab:v3-results}
\begin{tabular}{@{}lllll@{}}
\toprule
Contrast & W/L/T & Median $\Delta$ & Sign $p$ & Wilcoxon $p$ \\
\midrule
\multicolumn{5}{@{}l}{\emph{Continued pretraining}} \\
MLM+JEPA vs.\ step-matched MLM-only    & 76/15/23 & $+0.0051$ & $<$0.0001 & $<$0.0001 \\
MLM+JEPA vs.\ compute-matched MLM-only & 78/14/22 & $+0.0049$ & $<$0.0001 & $<$0.0001 \\
MLM+JEPA vs.\ off-the-shelf checkpoint & 81/20/13 & $+0.0068$ & $<$0.0001 & $<$0.0001 \\
Step-matched MLM-only vs.\ off-the-shelf checkpoint & 52/35/27 & $+0.0012$ & 0.043 & 0.15 \\
Ladder (extra shallow targets) vs.\ single half-depth target & 11/15/10 & $-0.0003$ & 0.84 & 0.82 \\
\midrule
\multicolumn{5}{@{}l}{\emph{From scratch (random init)}} \\
MLM+JEPA vs.\ step-matched MLM-only    & 50/41/23 & $+0.0013$ & 0.20 & 0.036 \\
MLM+JEPA vs.\ compute-matched MLM-only & 54/38/22 & $+0.0013$ & 0.059 & 0.035 \\
\bottomrule
\end{tabular}
\end{table}

\paragraph{Improvement over the pretrained checkpoint.}
MLM+JEPA is 81/20/13 against the off-the-shelf ESM2 checkpoint ($p{<}0.0001$,
median $+0.0068$), while MLM-only continuation is 52/35/27 with a median of
only $+0.0012$. At 150M, for example, SCOPe-40 Recall@1 is
0.424 off the shelf, 0.404 after compute-matched MLM-only continuation, and
0.507 with MLM+JEPA.

\subsection{The largest gains are on structure- and homology-sensitive tasks}
\label{sec:structure}

On the structure/homology subset defined in Sec.~\ref{sec:setup},
the median gain over step-matched MLM-only across both scales and seeds is
$+0.0132$ versus $+0.0045$ on the remaining tasks, roughly a threefold difference. Against the
compute-matched control the corresponding medians are $+0.0106$ and $+0.0041$
(Fig.~\ref{fig:v3-pertask}). Most of the subset gain comes from the two fold
tasks: across the six runs the median gain over the compute-matched control is
$+0.0614$ on SCOPe-40 retrieval and $+0.0267$ on remote homology, while
molecular function gains little but in every run ($+0.0037$) and subcellular
localization does not change ($+0.0008$). The next-largest gains are outside
the subset: peptide--HLA binding ($+0.0169$), signal-peptide residues
($+0.0137$), fluorescence ($+0.0124$), and $\beta$-lactamase ($+0.0117$), each
positive in at least five of six runs. The subset gain is larger at 150M than at 35M:
median $+0.026$ versus $+0.008$ (step-matched, four tasks, three seeds).
Against the compute-matched control, SCOPe-40 gains $+0.015$, $+0.044$, and $+0.103$ at 8M, 35M, and
150M, and remote homology $-0.008$, $+0.013$, and $+0.034$; the median over
the other tasks does not grow ($+0.007$, $+0.007$, $+0.003$;
Appendix~\ref{app:scale-trend}). The gain concentrates with scale rather
than spreading.

Five of 19 tasks beat the compute-matched control beyond the tie band in all
six scale$\times$seed runs: SCOPe-40 retrieval, remote homology, molecular
function, peptide--HLA binding, and signal-peptide residues. No task loses in
all six. On the three residue-level tasks, the continued JEPA arms of every controlled
backbone (ESM2-8M, 35M, 150M, and ProGen3) beat the compute-matched control in
19 of 24 comparisons (Table~\ref{tab:residue}); this gain is largest at 8M and
shrinks with size (Appendix~\ref{app:scale-trend}).

The SCOPe-40 gain changes little when retrieval is
scored after mean-centering or, in the wall-clock study, with Euclidean or
centered-and-whitened cosine distance
(Appendix~\ref{app:controlled-robustness}); the gain therefore persists beyond the
cosine similarity used in training.

\subsection{No reliable gain from random initialization}
\label{sec:scratch}

MLM+JEPA does not produce a reliable improvement across the 19 tasks from
random weights. Pooled over the six scale$\times$seed runs, it is 50/41/23 against the
step-matched control and 54/38/22 against the compute-matched one (sign test
$p{=}0.20$ and $0.059$; Wilcoxon $0.036$ and $0.035$), and the result does not
replicate across seeds (Appendix~\ref{app:controlled-robustness}). The
structure/homology subset trends in the same direction as continued
pretraining (median $+0.0059$ versus $+0.0132$ continued, step-matched;
$p{=}0.021$ uncorrected).

\subsection{Additional shallow targets reduce the structural advantage}
\label{sec:ladder}

The ladder ablation keeps the half-depth target and adds shallower targets.
Against the half-depth target, the ladder records 11 wins, 15 losses,
and 10 ties across 18 tasks at two scales ($p{=}0.84$).
It loses all eight comparisons on the structure/homology subset, reducing the
median gain over step-matched MLM-only from $+0.013$ to $+0.004$.

\FloatBarrier
\subsection{The Wall-clock study}
\label{sec:breadth}

\begin{figure}[!htb]
\centering
\includegraphics[width=0.9\linewidth]{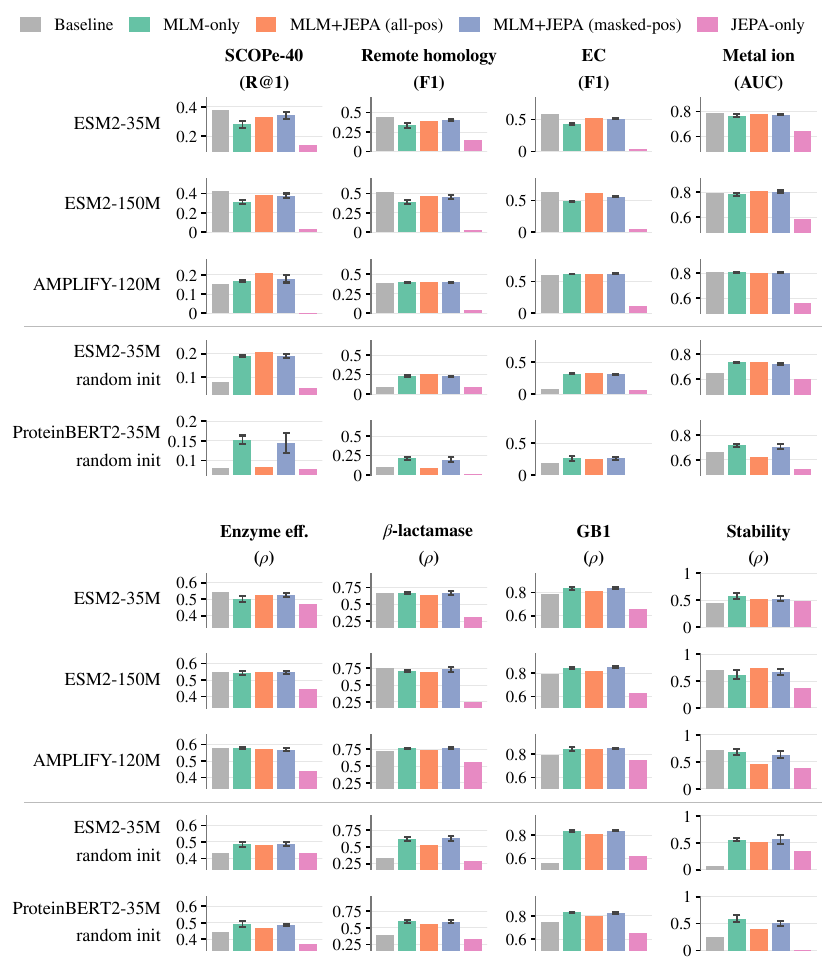}
\caption{Wall-clock study: eight tasks across five backbones.
Error bars: $\pm$1 standard deviation across pretraining seeds for MLM-only and
masked-position MLM+JEPA; other conditions use one run or checkpoint.
Baseline: pretrained checkpoint or untrained model.
R@1: Recall@1; $\rho$: Spearman; F1: macro-F1.}
\label{fig:wallclock-breadth}
\end{figure}

The pretrained ESM2 backbones favor MLM+JEPA (10/4/5 at 35M and 12/4/3 at
150M), AMPLIFY-120M is near parity (6/5/8), and the two random-init backbones
are at parity or worse (6/8/5 for ESM2-35M, 4/10/5 for ProteinBERT2-35M);
Fig.~\ref{fig:wallclock-breadth} shows eight of the 19 tasks and Appendix
Table~\ref{tab:masked-pos-headline} all of them. Across the 95
task-by-backbone comparisons,
ten differences survive Benjamini--Hochberg correction, all positive: SCOPe-40
retrieval, remote homology, and EC classification at both ESM2 scales; enzyme
catalytic efficiency, metal-ion binding, and secondary structure at 35M; and
disorder on random-init ESM2-35M. Five remain
significant under the stricter Holm correction; no negative comparison survives
either. On pretrained ESM2 both trained arms end 8\,h below the checkpoint on the
macro-mean, MLM+JEPA by less (Appendix Table~\ref{tab:headline-summary}).
Under wall-clock matching, MLM+JEPA gains occur despite fewer optimizer steps
and token exposures than MLM-only (Table~\ref{tab:training-coverage-appendix}).

\FloatBarrier
\subsection{Next-token prediction shows the same pattern}
\label{sec:ntp}

To test whether the gain extends beyond masked language modeling, we adapt the
controlled recipe to ProGen3-112m, a causal mixture-of-experts model trained
by next-token prediction (NTP). The latent loss predicts the teacher's
half-depth state four positions ahead of the student rather than at masked
positions (Appendix~\ref{app:ntp}). On the 16 tasks scored for this model, continued
NTP+JEPA beats step-matched NTP-only on 11 (4 losses) and compute-matched
NTP-only on 12 (3 losses; $p{=}0.018$). Wins include SCOPe-40 retrieval, remote homology, subcellular
localization, and all three residue-level tasks (Table~\ref{tab:ntp-results});
token-prediction loss does not improve (Appendix~\ref{app:ntp}).
Both arms improve over the checkpoint, with median gains of $+0.017$ for NTP+JEPA
and $+0.007$ for NTP-only. From random initialization, NTP+JEPA scores 10/3/3
against either control (sign-test $p{=}0.046$).

\begin{table}[!htbp]
\centering
\footnotesize
\setlength{\tabcolsep}{4pt}
\caption{Next-token prediction on ProGen3-112m, 16 tasks. Notation as in Table~\ref{tab:v3-results}; compute-matched NTP-only receives 14{,}070 extra steps, matching NTP+JEPA's 21\% overhead; per-task scores in Appendix~\ref{app:ntp}.}
\label{tab:ntp-results}
\begin{tabular}{@{}lllll@{}}
\toprule
Contrast & W/L/T & Median $\Delta$ & Sign $p$ & Wilcoxon $p$ \\
\midrule
\multicolumn{5}{@{}l}{\emph{Continued pretraining}} \\
NTP+JEPA vs.\ step-matched NTP-only & 11/4/1 & $+0.0072$ & 0.059 & 0.017 \\
NTP+JEPA vs.\ compute-matched NTP-only & 12/3/1 & $+0.0084$ & 0.018 & 0.0026 \\
NTP+JEPA vs.\ off-the-shelf checkpoint & 13/1/2 & $+0.0169$ & 0.00092 & 0.0017 \\
Step-matched NTP-only vs.\ off-the-shelf checkpoint & 12/3/1 & $+0.0073$ & 0.018 & 0.0055 \\
\midrule
\multicolumn{5}{@{}l}{\emph{From scratch (random init)}} \\
NTP+JEPA vs.\ step-matched NTP-only & 10/3/3 & $+0.0061$ & 0.046 & 0.088 \\
NTP+JEPA vs.\ compute-matched NTP-only & 10/3/3 & $+0.0056$ & 0.046 & 0.058 \\
\bottomrule
\end{tabular}
\end{table}

\FloatBarrier
\subsection{Loss form, target positions, and removing MLM}
\label{sec:objective-ablations}

We ablate three recipe choices: the form of the latent loss, the positions it
supervises, and whether MLM is needed at all. A fixed-step ESM2-35M factorial
with the wall-clock target and regularizer
(Table~\ref{tab:two-studies}) crossed masked-position versus
all-position latent targets with cosine, mean squared error (MSE), and
SmoothL1 losses at 25{,}000 steps
(Appendix Table~\ref{tab:loss-form-ablation}). The loss form, not the choice of
target positions, is what matters: masked-position and all-position targets
are indistinguishable (14/16/18, $p{=}0.856$, two-sided), whereas cosine gives
the highest mean downstream scores and beats MSE (16/5/11, $p{=}0.027$); its
difference from SmoothL1 is not significant.

JEPA-only causes large downstream losses on every backbone in the
wall-clock study (Appendix Table~\ref{tab:headline-summary}) and on ESM2-35M
with the controlled recipe (Appendix~\ref{app:controlled-robustness}), so
latent prediction complements rather than replaces token prediction.

\FloatBarrier
\section{Discussion and Limitations}

Latent prediction improves pretrained models' downstream performance in
every seed and at every size tested while its token-prediction loss is
0.002--0.005 nats worse (Sec.~\ref{sec:results-continued}), i.e., language
modeling loss does not necessarily track downstream quality in Biology. The concentration of gains on
SCOPe-40 retrieval and remote homology supports the hypothesis that
intermediate-layer targets help retain structural information; direct
layer-wise measurements are needed to establish that mechanism. This echoes
observations that some biological tasks do not improve with model or pretraining data scale,
while structure-related protein tasks
do~\citep{brandes_proteinbert_2022,breton_plm_2026,leib_dnt_2026,li_feature_2024}.

The size trend points the same way. Larger checkpoints hold more structural
information to begin with (off-the-shelf SCOPe-40 Recall@1 rises from 0.275
at 8M to 0.424 at 150M), continued MLM-only training loses more of it at the
larger sizes, and the latent gain over the compute-matched control grows in
step, two- to threefold with each roughly fourfold increase in parameters,
while median gain on the other tasks does not grow
(Table~\ref{tab:scale-trend}). Latent prediction therefore adds most where
the model already has a strong representation, and adds little from random
initialization, where there is no meaningful representation to update.
Three sizes do not establish a scaling law, but the monotone trend suggests
one; whether it continues past 150M, and whether more data or larger models
close the from-scratch gap, are open.

The controlled recipe ran one epoch over UniRef50 and
frozen-representation evaluation only; the wall-clock study broadens the
architecture coverage under wall-clock matching alone and UniRef30. We did not test larger models or datasets (such as UniRef100). The EMA teacher and the VISReg regularizer
were adopted together; neither was ablated on its own.

\section{Conclusion}

Latent prediction improves protein language model pretraining when combined
with token prediction. The gain holds under both step and compute matching,
reproduces across seeds and model sizes, transfers to a causal model, and is
largest on structure- and homology-sensitive tasks, where it grows with scale.
At 17\% extra compute it returns more than the same compute spent on pure
MLM. The recipe uses a half-depth EMA-teacher target with a cosine loss at masked positions. Ablations show that cosine beats mean squared error, adding shallower targets reduces structural gains, and removing MLM harms task performance. ProteinJEPA therefore complements rather than
replaces MLM and NTP, and is most useful for improving a pretrained model.

\section*{Ethics Statement}

This work changes a self-supervised training objective and evaluates
representations on public benchmarks. As with
other biological models, it may support both
beneficial and dual-use applications.

\section*{Reproducibility Statement}

The anonymized repository \url{https://anonymous.4open.science/r/protJepa-FF24} contains code, benchmark harness, and analysis scripts. Recipes, pretraining
seeds, matching criteria, and per-task metrics are documented in
Secs.~\ref{sec:method}--\ref{sec:experiments} and the appendix. Model checkpoints will be
released after de-anonymization.

\section*{AI Use Statement}

Generative AI tools assisted with software development, 
result interpretation, and manuscript proofing. The authors reviewed the code, analyses, claims, prose and take responsibility for the final content.

\bibliographystyle{iclr2027_conference}
\bibliography{references}

\newpage

\FloatBarrier
\appendix
\raggedbottom
\section{Additional Results and Reproducibility}

\subsection{Controlled study: robustness of the headline contrast}
\label{app:controlled-robustness}

The continued-pretraining result holds under changes to the tie band and
scale pooling. Alternative metrics preserve the pooled contrasts but weaken
some per-scale tests.

\paragraph{Tie band.}
The table below compares the $|\Delta|<0.002$ tie band used in
Table~\ref{tab:v3-results} with exact ties only. Within each contrast, the
three rows are the three pretraining seeds in a fixed order:

\begin{center}
\footnotesize
\renewcommand{\arraystretch}{1.12}
\setlength{\tabcolsep}{5pt}
\begin{tabular}{@{}lllll@{}}
\toprule
 & \multicolumn{2}{c}{Band $0.002$} & \multicolumn{2}{c}{Exact ties only} \\
\cmidrule(lr){2-3}\cmidrule(l){4-5}
Contrast & W/L/T & $p$ & W/L/T & $p$ \\
\midrule
Continued vs.\ step-matched & 26/5/7 & 0.0001 & 28/10/0 & 0.0025 \\
 & 25/6/7 & 0.0004 & 29/9/0 & 0.0008 \\
 & 25/4/9 & 0.0001 & 31/7/0 & 0.0001 \\
\addlinespace
Continued vs.\ compute-matched & 27/5/6 & 0.0001 & 30/8/0 & 0.0002 \\
 & 26/5/7 & 0.0001 & 30/8/0 & 0.0002 \\
 & 25/4/9 & 0.0001 & 32/6/0 & $<$0.0001 \\
\addlinespace
From scratch vs.\ step-matched & 15/17/6 & 0.70 & 17/21/0 & 0.79 \\
 & 17/10/11 & 0.12 & 25/13/0 & 0.037 \\
 & 18/14/6 & 0.30 & 20/18/0 & 0.44 \\
\addlinespace
From scratch vs.\ compute-matched & 16/16/6 & 0.57 & 18/20/0 & 0.69 \\
 & 20/9/9 & 0.031 & 23/15/0 & 0.13 \\
 & 18/13/7 & 0.24 & 21/17/0 & 0.31 \\
\midrule
From scratch, pooled              & 50/41/23 & 0.20 & 62/52/0 & 0.20 \\
\bottomrule
\end{tabular}
\end{center}

\paragraph{Pooling the two scales.}
Table~\ref{tab:v3-results} pools 19 tasks over two scales and three seeds. Because the scales share a
task list, we also report two estimators that do not treat scale--task pairs as
independent. The per-task means are reported per seed in the same order; the
single-scale rows use the first seed:

\begin{center}
\footnotesize
\renewcommand{\arraystretch}{1.12}
\setlength{\tabcolsep}{5pt}
\begin{tabular}{@{}llll@{}}
\toprule
Contrast (continued) & Estimator & W/L/T & Sign $p$ \\
\midrule
vs.\ step-matched & per-task mean over scales ($n{=}19$) & 14/1/4 & 0.0005 \\
 & & 13/2/4 & 0.0037 \\
 & & 13/1/5 & 0.0009 \\
\addlinespace
vs.\ compute-matched & per-task mean over scales ($n{=}19$) & 14/1/4 & 0.0005 \\
 & & 13/2/4 & 0.0037 \\
 & & 14/2/3 & 0.0021 \\
\midrule
vs.\ step-matched    & 35M alone ($n{=}19$)  & 13/2/4 & 0.0037 \\
vs.\ step-matched    & 150M alone ($n{=}19$) & 13/3/3 & 0.011 \\
\bottomrule
\end{tabular}
\end{center}

\paragraph{Retrieval metric.}
Mean-centering SCOPe-40 embeddings leaves the controlled-study gain nearly
unchanged: $+0.0431 \rightarrow +0.0403$ at 35M and
$+0.1052 \rightarrow +0.1001$ at 150M. In the wall-clock study, Euclidean and
centered-and-whitened cosine scoring move the masked-position gain by less than
one percentage point.

\paragraph{Alternative evaluation metrics.}
The wall-clock study fixed a different metric for nine shared tasks
(Table~\ref{tab:task-sources}). Using those metrics instead, the pooled and
per-task-averaged headline contrasts are unchanged. Per-scale contrasts
are more sensitive and can lose significance at $\alpha{=}0.05$ under those
metrics.

\begin{figure}[H]
\centering
\includegraphics[width=\linewidth]{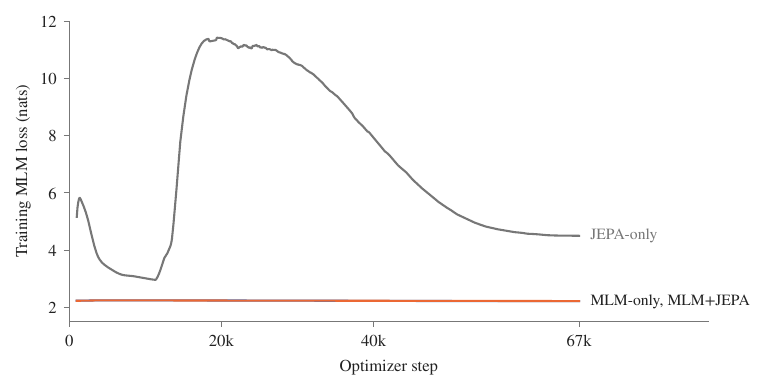}
\caption{Training MLM loss with JEPA-only continuation (ESM2-35M; MLM weight
0), monitored but not optimized. MLM-only and MLM+JEPA coincide at this scale.
JEPA-only reaches 11.4 nats by step 20{,}000 and settles near 4.5.}
\label{fig:loss-jepa-only}
\end{figure}

\paragraph{JEPA-only at the controlled recipe.}
The JEPA-only evidence elsewhere in this appendix comes from the wall-clock
study, whose runs used an all-position target and different regularization
(Table~\ref{tab:recipe-sweep-app}). Repeating the ablation at the controlled
recipe --- masked-position cosine with the EMA teacher, $15\%$ masking, 67{,}000
steps, continued from pretrained ESM2-35M with the MLM weight set to
zero --- also degrades downstream performance. Against the matched MLM-only
control, W/L/T is 2/17/0 with median $\Delta=-0.124$ ($n=19$), and against the
off-the-shelf checkpoint it is 2/17/0 with median $\Delta=-0.129$. The largest
losses are EC ($-0.481$), remote homology ($-0.357$), and molecular function
($-0.320$). Its training MLM loss, monitored but not optimized, climbs to 11.4
nats before settling near 4.5 (Fig.~\ref{fig:loss-jepa-only}).

\begin{figure}[!htbp]
\centering
\includegraphics[width=\linewidth]{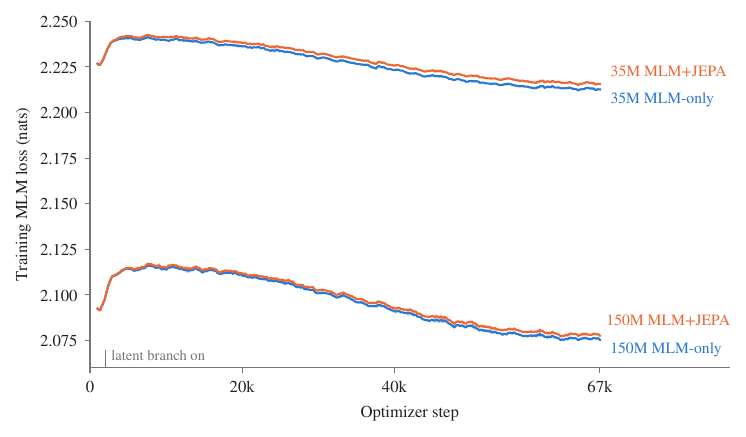}
\caption{Training MLM loss during continued pretraining (1{,}000-step moving
average; line: mean over the two seeds with complete 67{,}000-step logs;
shading: seed range).
Paired arms see the same batches. MLM+JEPA's loss is about 0.002 nats higher
at either scale. The latent branch switches on at step 2{,}000.}
\label{fig:loss-curves}
\end{figure}

\begin{figure}[!htbp]
\centering
\includegraphics[width=\linewidth]{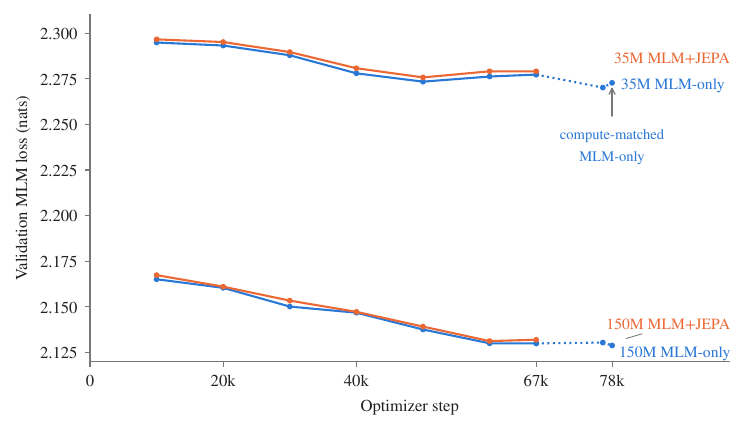}
\caption{Validation MLM loss (one pretraining seed; 20{,}000 held-out UniRef50 sequences, every
10{,}000 steps). Dotted: the compute-matched MLM-only control, 11{,}390 further
steps at the final learning rate, changing loss by at most 0.005 nats.
MLM+JEPA's loss is 0.001--0.003 nats higher than MLM-only's.}
\label{fig:val-loss}
\end{figure}

\FloatBarrier

\FloatBarrier

\subsection{Model-size trend: an ESM2-8M point}
\label{app:scale-trend}

\begin{figure}[H]
\centering
\includegraphics[width=\linewidth]{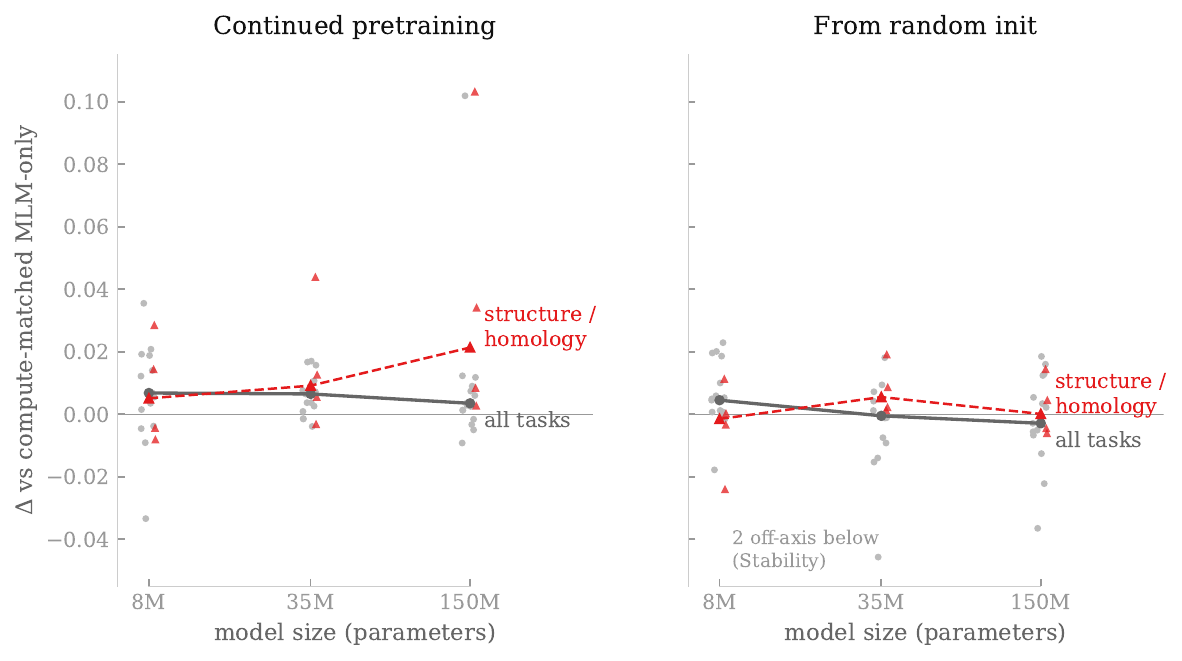}
\caption{MLM+JEPA minus compute-matched MLM-only against model size (one seed
at every size; 19 tasks). Dots are tasks, lines are
medians; triangles mark the structure/homology subset. Stability values below
$-0.055$ are clipped to the axis. The subset median rises with size because SCOPe-40 and
remote homology do; the other tasks do not.}
\label{fig:scale-trend}
\end{figure}

\begin{figure}[H]
\centering
\includegraphics[width=0.85\linewidth]{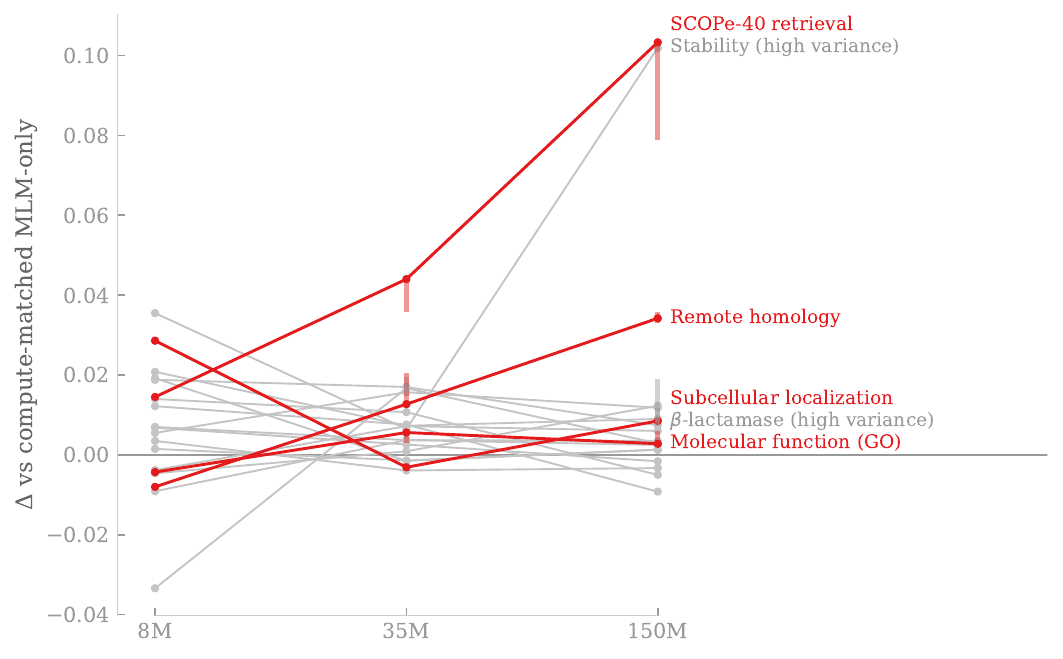}
\caption{Per-task deltas across sizes, continued pretraining. Vertical ticks
at 35M and 150M are three-seed ranges for the five tasks that beat the control
in every seed (Sec.~\ref{sec:structure}); the 8M SCOPe-40 and remote-homology
points fall below the 35M range.}
\label{fig:scale-slope}
\end{figure}

To see how the continued-pretraining gain changes with model size we ran the
controlled recipe unchanged on ESM2-8M (6 layers, half-depth target at
layer 3), including a compute-matched MLM-only control
whose 12{,}080 extra steps come from the 18\% overhead measured at this size.
Table~\ref{tab:scale-trend} and Figs.~\ref{fig:scale-trend}--\ref{fig:scale-slope}
compare the three sizes on all 19 tasks; the 35M and 150M entries use one seed
so that every size is a single run, and their three-seed ranges appear in
Fig.~\ref{fig:scale-slope}. SCOPe-40 retrieval and remote homology gain more
at each step in size, and the 8M point lies below the 35M seed range. The
other 17 tasks, including molecular function and subcellular localization,
show no such trend. From random initialization no task trends.
The three residue-level tasks (Table~\ref{tab:residue}) run the other way:
their median gain over the compute-matched control falls from $+0.012$ at 8M
to $+0.008$ at 35M and $+0.004$ at 150M.
Table~\ref{tab:8m-eval-loss} lists the 8M held-out losses: MLM+JEPA is
0.003--0.005 nats worse on MLM loss at both initializations, as at 35M and
150M.

\begin{table}[H]
\centering
\scriptsize
\setlength{\tabcolsep}{4pt}
\caption{Model-size trend on the 19 tasks. $\Delta$ is MLM+JEPA minus compute-matched MLM-only, one pretraining seed at every size; brackets give the three-seed range at 35M and 150M. The last three rows are absolute SCOPe-40 Recall@1.}
\label{tab:scale-trend}
\begin{tabular}{@{}llll@{}}
\toprule
 & ESM2-8M & ESM2-35M & ESM2-150M \\
\midrule
\multicolumn{4}{@{}l}{\emph{Continued pretraining}} \\
$\Delta$ SCOPe-40 retrieval & $+0.015$ & $+0.044$ [+0.036, +0.044] & $+0.103$ [+0.079, +0.103] \\
$\Delta$ remote homology & $-0.008$ & $+0.013$ [+0.013, +0.020] & $+0.034$ [+0.033, +0.036] \\
median $\Delta$, structure/homology (4 tasks) & $+0.005$ & $+0.009$ & $+0.021$ \\
median $\Delta$, other 15 tasks & $+0.007$ & $+0.007$ & $+0.003$ \\
W/L/T, all 19 tasks & 12/6/1 & 14/2/3 & 13/3/3 \\
\addlinespace
SCOPe-40 Recall@1, off-the-shelf checkpoint & 0.275 & 0.383 & 0.424 \\
SCOPe-40 Recall@1, compute-matched MLM-only & 0.279 & 0.368 & 0.404 \\
SCOPe-40 Recall@1, MLM+JEPA & 0.293 & 0.412 & 0.507 \\
\midrule
\multicolumn{4}{@{}l}{\emph{From random init}} \\
$\Delta$ SCOPe-40 retrieval & $+0.011$ & $+0.001$ & $+0.015$ \\
median $\Delta$, structure/homology (4 tasks) & $-0.002$ & $+0.006$ & $+0.000$ \\
median $\Delta$, other 15 tasks & $+0.005$ & $-0.001$ & $-0.003$ \\
W/L/T, all 19 tasks & 10/3/6 & 7/6/6 & 9/10/0 \\
\bottomrule
\end{tabular}
\end{table}

\begin{table}[H]
\centering
\footnotesize
\caption{ESM2-8M held-out losses at the end of training (6{,}810 UniRef50 validation sequences). MLM loss is masked-token cross-entropy in nats; the JEPA loss is one minus cosine similarity at masked positions and is defined only for the MLM+JEPA arms. Compute-matched arms report the end of their 12{,}080 extra steps.}
\label{tab:8m-eval-loss}
\begin{tabular}{@{}lrrr@{}}
\toprule
Arm & Step & Eval MLM loss & Eval JEPA loss \\
\midrule
\multicolumn{4}{@{}l}{\emph{Continued pretraining}} \\
MLM-only, step-matched & 67,000 & 2.391 & --- \\
MLM-only, compute-matched & 12,080 & 2.392 & --- \\
MLM+JEPA & 67,000 & 2.396 & 0.072 \\
\midrule
\multicolumn{4}{@{}l}{\emph{From random init}} \\
MLM-only, step-matched & 67,000 & 2.445 & --- \\
MLM-only, compute-matched & 12,080 & 2.443 & --- \\
MLM+JEPA & 67,000 & 2.448 & 0.051 \\
\bottomrule
\end{tabular}
\end{table}

\FloatBarrier

\subsection{Per-residue probes}
\label{app:residue}

The three residue-level tasks are \emph{eight-state secondary structure} (DSSP labels from
NetSurfP-2.0, macro-F1), \emph{intrinsic disorder} (the NetSurfP-2.0 disorder
mask, Matthews correlation coefficient, MCC), and \emph{signal peptide} (SignalP~6.0 per-residue signal-peptide
labels, macro-F1). Each is a linear probe (a linear head on frozen final-layer
residue embeddings), trained on the task's training split and scored on its
test split. Table~\ref{tab:residue} breaks their JEPA-minus-control
differences down by backbone, including ESM2-8M and ProGen3. Under the
controlled recipe, the continued JEPA arms beat the step-matched control in 20 of 24 run-by-task comparisons,
the compute-matched control in 19, and the off-the-shelf checkpoint in 21
(sign test $p{<}0.001$ each); from random initialization it is 12/4/8 against
the compute-matched control ($p{=}0.038$). In the wall-clock study the contrast
against MLM-only is 21/8/1 on the two pretrained ESM2 backbones ($p{=}0.012$)
and 41/28/9 over all five ($p{=}0.074$).

\begin{table}[H]
\centering
\footnotesize
\caption{Per-residue probes: MLM+JEPA (NTP+JEPA for ProGen3) minus its control on three residue-level tasks (eight-state secondary structure, intrinsic disorder, signal peptide), pooled over pretraining runs and tasks. Controlled study: per-backbone rows are against the compute-matched token-prediction control; the last rows of each block pool all four backbones against each control. Wall-clock study: against MLM-only, paired by pretraining seed. Notation as in Table~\ref{tab:v3-results}; $p$-values are one-sided.}
\label{tab:residue}
\begin{tabular}{@{}llllll@{}}
\toprule
Backbone & $n$ & W/L/T & Median $\Delta$ & Sign $p$ & Wilcoxon $p$ \\
\midrule
\multicolumn{6}{@{}l}{\emph{Controlled study, continued pretraining}} \\
ESM2-8M & 3 & 3/0/0 & $+0.0121$ & 0.125 & 0.125 \\
ESM2-35M & 9 & 8/0/1 & $+0.0076$ & 0.004 & 0.004 \\
ESM2-150M & 9 & 5/2/2 & $+0.0041$ & 0.227 & 0.150 \\
ProGen3-112m & 3 & 3/0/0 & $+0.0082$ & 0.125 & 0.125 \\
All, vs compute-matched & 24 & 19/2/3 & $+0.0079$ & $<$0.001 & $<$0.001 \\
All, vs step-matched & 24 & 20/2/2 & $+0.0074$ & $<$0.001 & $<$0.001 \\
All, vs off-the-shelf checkpoint & 24 & 21/2/1 & $+0.0112$ & $<$0.001 & $<$0.001 \\
\midrule
\multicolumn{6}{@{}l}{\emph{Controlled study, from random init}} \\
ESM2-8M & 3 & 2/0/1 & $+0.0045$ & 0.250 & 0.250 \\
ESM2-35M & 9 & 3/2/4 & $+0.0003$ & 0.500 & 0.285 \\
ESM2-150M & 9 & 5/2/2 & $+0.0021$ & 0.227 & 0.125 \\
ProGen3-112m & 3 & 2/0/1 & $+0.0046$ & 0.250 & 0.125 \\
All, vs compute-matched & 24 & 12/4/8 & $+0.0020$ & 0.038 & 0.008 \\
All, vs step-matched & 24 & 10/5/9 & $+0.0016$ & 0.151 & 0.028 \\
\midrule
\multicolumn{6}{@{}l}{\emph{Wall-clock study, 8\,h, vs MLM-only}} \\
ESM2-35M & 15 & 11/4/0 & $+0.0074$ & 0.059 & 0.018 \\
ESM2-150M & 15 & 10/4/1 & $+0.0031$ & 0.090 & 0.084 \\
AMPLIFY-120M & 18 & 7/7/4 & $+0.0009$ & 0.605 & 0.695 \\
ESM2-35M (random init) & 15 & 7/6/2 & $+0.0010$ & 0.500 & 0.423 \\
ProteinBERT2-35M & 15 & 6/7/2 & $-0.0014$ & 0.709 & 0.820 \\
Pretrained ESM2 (35M and 150M) & 30 & 21/8/1 & $+0.0040$ & 0.012 & 0.004 \\
All & 78 & 41/28/9 & $+0.0022$ & 0.074 & 0.116 \\
\bottomrule
\end{tabular}
\end{table}

\FloatBarrier

\subsection{Wall-clock study: the full objective grid}
\label{app:allpos}

This section reports the wall-clock study's two remaining objectives --- the
all-position variant and JEPA-only --- across the full
5\,$\times$\,3 matrix of five backbones and three budgets.

\begin{table}[H]
\centering
\scriptsize
\caption{Wall-clock macro-mean linear-probe $\Delta$ relative to each backbone's
baseline: 16 protein-level tasks at 8\,h, 15 (no EC) at 1\,h and 4\,h. All-pos MLM+JEPA is the all-position MSE control;
Masked-pos MLM+JEPA is the masked-position cosine recipe at 8\,h.
$\Delta\Delta$ is All-pos MLM+JEPA
minus MLM-only; a Holm-corrected paired Wilcoxon test finds no rejection
(smallest $p_{H}=0.15$). Masked-pos pools pretraining seeds; other trained
columns are single runs. Table~\ref{tab:loss-form-ablation} separates target
positions from loss form at fixed steps.}
\label{tab:headline-summary}
\begin{tabular}{@{}llccccc@{}}
\toprule
Backbone & Time & MLM-only & \shortstack{All-pos\\MLM+JEPA} & $\Delta\Delta$ & JEPA-only & \shortstack{Masked-pos\\MLM+JEPA} \\
\midrule
ESM2-35M & 1h & $-0.011$ & $-0.012$ & $-0.001$ & $-0.235$ & -- \\
 & 4h & $-0.014$ & $-0.007$ & $+0.007$ & $-0.222$ & -- \\
 & 8h & $-0.014$ & $-0.005$ & $+0.010$ & $-0.130$ & $-0.001$ \\
\midrule
ESM2-150M & 1h & $-0.007$ & $-0.004$ & $+0.003$ & $-0.208$ & -- \\
 & 4h & $-0.016$ & $-0.003$ & $+0.013$ & $-0.248$ & -- \\
 & 8h & $-0.028$ & $-0.009$ & $+0.019$ & $-0.246$ & $-0.013$ \\
\midrule
AMPLIFY-120M & 1h & $-0.004$ & $-0.040$ & $-0.036$ & $-0.181$ & -- \\
 & 4h & $-0.005$ & $-0.012$ & $-0.007$ & $-0.239$ & -- \\
 & 8h & $-0.005$ & $-0.015$ & $-0.010$ & $-0.201$ & $+0.004$ \\
\midrule
ESM2-35M (random init) & 1h & $+0.125$ & $+0.121$ & $-0.004$ & $-0.032$ & -- \\
 & 4h & $+0.140$ & $+0.139$ & $-0.001$ & $+0.036$ & -- \\
 & 8h & $+0.151$ & $+0.140$ & $-0.011$ & $+0.015$ & $+0.149$ \\
\midrule
ProteinBERT2-35M & 1h & $-0.007$ & $+0.003$ & $+0.010$ & $-0.076$ & -- \\
 & 4h & $+0.032$ & $+0.028$ & $-0.005$ & $-0.112$ & -- \\
 & 8h & $+0.012$ & $+0.050$ & $+0.038$ & $-0.127$ & $+0.098$ \\
\bottomrule
\end{tabular}
\end{table}

\paragraph{All-position MLM+JEPA stays close to MLM-only.}
The macro-mean gap between all-position MLM+JEPA and MLM-only is
$|\Delta\Delta|\le 0.041$ across all 15 backbone--budget pairs and $\le 0.036$ on
pretrained backbones (Table~\ref{tab:headline-summary}). No difference is significant
after Holm correction. The masked-position recipe wins most matched
8\,h (backbone, task) comparisons against this control
($p\approx2\times10^{-4}$), but also changes MSE to cosine; the fixed-step
factorial (Table~\ref{tab:loss-form-ablation}) separates these choices.

\FloatBarrier

\subsection{Loss-form and target-position ablation}

\begin{table}[H]
\centering
\small
\caption{Fixed-step ESM2-35M factorial at 25{,}000 steps with the wall-clock target and regularizer: initialization $\times$ target positions $\times$ loss form. $N$ counts task--configuration comparisons, pooled as indicated; each score averages validation and test results. Ties are $|\Delta|\le0.002$. Reported $p$-values are two-sided; \textbf{bold} marks $p<0.05$.}
\label{tab:loss-form-ablation}
\begin{tabular}{@{}lrrrrrrrr@{}}
\toprule
Comparison & $N$ & W & L & T & Mean $\Delta$ & $p_{\text{sign}}$ & $p_{\text{Wilcox}}$ & $p_t$ \\
\midrule
\multicolumn{9}{@{}l}{\textit{Loss form (pretrained, positions pooled)}}\\
\hspace{1em}cosine vs.\ MSE & 32 & 16 & 5 & 11 & $+0.005$ & $\mathbf{0.027}$ & $\mathbf{0.019}$ & $\mathbf{0.026}$ \\
\hspace{1em}cosine vs.\ SmoothL1 & 32 & 10 & 8 & 14 & $+0.003$ & $0.815$ & $0.331$ & $0.130$ \\
\hspace{1em}MSE vs.\ SmoothL1 & 32 & 5 & 8 & 19 & $-0.002$ & $0.581$ & $0.295$ & $0.111$ \\
\midrule
\multicolumn{9}{@{}l}{\textit{Target set (pretrained, losses pooled)}}\\
\hspace{1em}masked vs.\ all-position & 48 & 14 & 16 & 18 & $+0.000$ & $0.856$ & $0.678$ & $0.670$ \\
\midrule
\multicolumn{9}{@{}l}{\textit{Initialization (all losses, all positions)}}\\
\hspace{1em}pretrained vs.\ random-init & 96 & 96 & 0 & 0 & $+0.361$ & $\mathbf{<\!10^{-15}}$ & $\mathbf{<\!10^{-15}}$ & $\mathbf{<\!10^{-15}}$ \\
\bottomrule
\end{tabular}
\end{table}

\subsection{All-position recipe sweep on pretrained ESM2-35M}

A matched-wall-clock sweep on pretrained ESM2-35M screened 37
all-position configurations over target construction, loss, predictor, and
regularization (selected rows in Table~\ref{tab:recipe-sweep-app}). We selected the
gradient-blocked target recipe because it needs no teacher network; a sub-sweep over $\lambda\in\{0.10,0.25,0.45,1.00\}$ selected $0.45$.
The masked-position cosine recipe was not in this sweep.

\begin{table}[!htbp]
\centering
\caption{All-position recipe-sweep results on continued ESM2-35M at matched
wall-clock. W/L/T are per task ($n=13$; 12 for JEPA-only). Mean $\Delta$ is relative to matched MLM-only.}
\label{tab:recipe-sweep-app}
\small
\begin{tabular}{@{}lcc@{}}
\toprule
Recipe & W/L/T & Mean $\Delta$ \\
\midrule
Baseline ESM2-35M (no continuation) & 5/7/1 & $+0.0026$ \\
JEPA + EMA teacher                 & 6/7/0 & $+0.0075$ \\
JEPA gradient-blocked target, $\lambda{=}0.45$ (selected) & 5/8/0 & $+0.0033$ \\
JEPA $\lambda{=}0.25$              & 5/7/1 & $-0.0029$ \\
JEPA + VICReg~\citep{bardes_vicreg_2022} & 4/9/0 & $-0.0020$ \\
JEPA $\lambda{=}0.10$              & 3/9/1 & $-0.0001$ \\
JEPA $\lambda{=}1.00$              & 3/9/1 & $-0.0020$ \\
ESM2 JEPA-only (matched)           & 2/9/1 & $-0.0095$ \\
\bottomrule
\end{tabular}
\end{table}

\FloatBarrier

\subsection{Training coverage and compute}
\label{app:training-coverage}
Table~\ref{tab:training-coverage-appendix} lists steps, samples, tokens, and
analytic FLOP estimates (sequence length 512) for the wall-clock checkpoints,
one run per arm.
FLOPs are approximate analytic estimates.
\begingroup
\scriptsize
\setlength{\tabcolsep}{1.5pt}
\setlength{\LTleft}{0pt}
\setlength{\LTright}{0pt}
\setlength{\LTcapwidth}{0.76\linewidth}
\begin{longtable}{@{}p{0.19\linewidth}p{0.175\linewidth}p{0.09\linewidth}rrrrr@{}}
\caption{\small Training coverage. Save: checkpoint hour. Units: millions of samples, billions of tokens, and $10^{18}$ FLOPs.}\label{tab:training-coverage-appendix}\\
\toprule
Model & Objective & Save & Steps & Batch & Samp. & Tok. & FLOPs \\
\midrule
\endfirsthead
\toprule
Model & Objective & Save & Steps & Batch & Samp. & Tok. & FLOPs \\
\midrule
\endhead
\midrule
\multicolumn{8}{r}{Continued on next page}\\
\midrule
\endfoot
\bottomrule
\endlastfoot
ESM2-35M & MLM-only & 1h & 19,633 & 128 & 2.51 & 1.29 & 0.258 \\
ESM2-35M & MLM-only & 4h & 80,107 & 128 & 10.25 & 5.25 & 1.054 \\
ESM2-35M & MLM-only & 8h & 160,793 & 128 & 20.58 & 10.54 & 2.116 \\
ESM2-35M & All-pos MLM+JEPA & 1h & 10,874 & 128 & 1.39 & 0.71 & 0.199 \\
ESM2-35M & All-pos MLM+JEPA & 4h & 43,993 & 128 & 5.63 & 2.88 & 0.805 \\
ESM2-35M & All-pos MLM+JEPA & 8h & 88,024 & 128 & 11.27 & 5.77 & 1.612 \\
ESM2-35M & JEPA-only & 1h & 10,569 & 128 & 1.35 & 0.69 & 0.193 \\
ESM2-35M & JEPA-only & 4h & 43,662 & 128 & 5.59 & 2.86 & 0.797 \\
ESM2-35M & JEPA-only & 8h & 87,929 & 128 & 11.25 & 5.76 & 1.604 \\
ESM2-35M & Masked-pos MLM+JEPA & 8h & 100,000 & 128 & 12.80 & 6.55 & 1.831 \\
ESM2-150M & MLM-only & 1h & 4,893 & 128 & 0.63 & 0.32 & 0.285 \\
ESM2-150M & MLM-only & 4h & 22,993 & 128 & 2.94 & 1.51 & 1.338 \\
ESM2-150M & MLM-only & 8h & 47,161 & 128 & 6.04 & 3.09 & 2.745 \\
ESM2-150M & All-pos MLM+JEPA & 1h & 1,673 & 128 & 0.21 & 0.11 & 0.132 \\
ESM2-150M & All-pos MLM+JEPA & 4h & 7,016 & 128 & 0.90 & 0.46 & 0.554 \\
ESM2-150M & All-pos MLM+JEPA & 8h & 14,433 & 128 & 1.85 & 0.95 & 1.140 \\
ESM2-150M & JEPA-only & 1h & 1,693 & 128 & 0.22 & 0.11 & 0.133 \\
ESM2-150M & JEPA-only & 4h & 7,135 & 128 & 0.91 & 0.47 & 0.563 \\
ESM2-150M & JEPA-only & 8h & 14,661 & 128 & 1.88 & 0.96 & 1.156 \\
ESM2-150M & Masked-pos MLM+JEPA & 8h & 44,894 & 48 & 2.15 & 1.10 & 1.329 \\
AMPLIFY-120M & MLM-only & 1h & 7,352 & 208 & 1.53 & 0.78 & 0.556 \\
AMPLIFY-120M & MLM-only & 4h & 28,165 & 208 & 5.86 & 3.00 & 2.129 \\
AMPLIFY-120M & MLM-only & 8h & 57,414 & 208 & 11.94 & 6.11 & 4.340 \\
AMPLIFY-120M & All-pos MLM+JEPA & 1h & 5,194 & 208 & 1.08 & 0.55 & 0.535 \\
AMPLIFY-120M & All-pos MLM+JEPA & 4h & 19,705 & 208 & 4.10 & 2.10 & 2.029 \\
AMPLIFY-120M & All-pos MLM+JEPA & 8h & 40,135 & 208 & 8.35 & 4.27 & 4.132 \\
AMPLIFY-120M & JEPA-only & 1h & 5,281 & 208 & 1.10 & 0.56 & 0.544 \\
AMPLIFY-120M & JEPA-only & 4h & 19,994 & 208 & 4.16 & 2.13 & 2.058 \\
AMPLIFY-120M & JEPA-only & 8h & 40,644 & 208 & 8.45 & 4.33 & 4.184 \\
ESM2-35M (random init) & MLM-only & 1h & 15,950 & 128 & 2.04 & 1.05 & 0.210 \\
ESM2-35M (random init) & MLM-only & 4h & 65,305 & 128 & 8.36 & 4.28 & 0.859 \\
ESM2-35M (random init) & MLM-only & 8h & 134,413 & 128 & 17.20 & 8.81 & 1.769 \\
ESM2-35M (random init) & All-pos MLM+JEPA & 1h & 9,514 & 128 & 1.22 & 0.62 & 0.174 \\
ESM2-35M (random init) & All-pos MLM+JEPA & 4h & 38,674 & 128 & 4.95 & 2.53 & 0.708 \\
ESM2-35M (random init) & All-pos MLM+JEPA & 8h & 78,818 & 128 & 10.09 & 5.17 & 1.443 \\
ESM2-35M (random init) & JEPA-only & 1h & 9,601 & 128 & 1.23 & 0.63 & 0.175 \\
ESM2-35M (random init) & JEPA-only & 4h & 38,945 & 128 & 4.98 & 2.55 & 0.710 \\
ESM2-35M (random init) & JEPA-only & 8h & 79,330 & 128 & 10.15 & 5.20 & 1.447 \\
ProteinBERT2-35M & MLM-only & 1h & 13,931 & 128 & 1.78 & 0.91 & 0.207 \\
ProteinBERT2-35M & MLM-only & 4h & 57,071 & 128 & 7.31 & 3.74 & 0.849 \\
ProteinBERT2-35M & MLM-only & 8h & 114,591 & 128 & 14.67 & 7.51 & 1.704 \\
ProteinBERT2-35M & All-pos MLM+JEPA & 1h & 9,167 & 128 & 1.17 & 0.60 & 0.190 \\
ProteinBERT2-35M & All-pos MLM+JEPA & 4h & 37,394 & 128 & 4.79 & 2.45 & 0.774 \\
ProteinBERT2-35M & All-pos MLM+JEPA & 8h & 75,019 & 128 & 9.60 & 4.92 & 1.552 \\
ProteinBERT2-35M & JEPA-only & 1h & 9,196 & 128 & 1.18 & 0.60 & 0.190 \\
ProteinBERT2-35M & JEPA-only & 4h & 37,521 & 128 & 4.80 & 2.46 & 0.776 \\
ProteinBERT2-35M & JEPA-only & 8h & 75,274 & 128 & 9.64 & 4.93 & 1.557 \\
ProteinBERT2-35M & Masked-pos MLM+JEPA & 8h & 64,256 & 128 & 8.22 & 4.21 & 1.329 \\
\end{longtable}
\endgroup

\subsection{Masked-position MLM+JEPA across backbones}
\label{app:maskedpos-extra-checkpoints}

Table~\ref{tab:masked-pos-headline} summarizes the per-backbone contrasts shown
in Fig.~\ref{fig:wallclock-breadth}.
\begin{table}[H]
\centering
\small
\caption{Masked-position MLM+JEPA versus matched MLM-only continuation after
8\,h on 19 tasks. W/L/T counts use $|\Delta|<0.002$ as the tie band. Median
$\Delta$ is relative to matched MLM-only and the sign-test $p$-value is
one-sided over wins and losses. Scores for both arms pool pretraining seeds.}
\label{tab:masked-pos-headline}
\begin{tabular}{@{}lccccc@{}}
\toprule
Backbone & W & L & T & Median $\Delta$ & Sign $p$ \\ 
\midrule
\multicolumn{6}{@{}l}{\textit{Pretrained continuation}}\\
\hspace{1em}ESM2-35M & 10 & 4 & 5 & $+0.004$ & 0.090 \\
\hspace{1em}ESM2-150M & 12 & 4 & 3 & $+0.006$ & 0.038 \\
\hspace{1em}AMPLIFY-120M & 6 & 5 & 8 & $-0.001$ & 0.500 \\
\midrule
\multicolumn{6}{@{}l}{\textit{Random init (from scratch)}}\\
\hspace{1em}ProteinBERT2-35M & 4 & 10 & 5 & $-0.002$ & 0.971 \\
\hspace{1em}ESM2-35M & 6 & 8 & 5 & $-0.001$ & 0.788 \\
\bottomrule
\end{tabular}
\end{table}

\FloatBarrier

\subsection{Benchmark tasks and sources}
\label{app:task-descriptions}

Table~\ref{tab:task-sources} lists the dataset identifier and the metric fixed
in advance for each task in each study. Both suites mix TAPE-derived tasks~\citep{rao_evaluating_2019} with
public benchmark datasets, plus the
ProteinBERT-style SignalP binary setup, the NeuroPID neuropeptide
benchmark~\citep{Karsenty2014,OferD2014}, and the SCOPe-40 retrieval benchmark.

The two suites are both 19 tasks and overlap in 18 of them: the wall-clock
study scores CheZoD disorder and the controlled study scores molecular
function (GO) in its place. Nine shared tasks carry different metrics, so their
scores are not comparable across studies.

\paragraph{Fold and function.}
\emph{Remote homology} predicts a protein's fold class; \emph{SCOPe-40}
retrieves a protein of the same SCOPe family, with Recall@1 measuring whether the
nearest retrieved sequence belongs to that family. \emph{Molecular function} predicts
Gene Ontology (GO) activity labels, while \emph{EC classification} predicts
Enzyme Commission reaction labels. \emph{Subcellular localization} predicts
the cellular compartment in which a protein resides \citep{thumuluri_deeploc_2022, gligorijevic_structure-based_2021}.

\paragraph{Binding and interactions.}
\emph{Peptide--HLA binding} classifies whether a peptide binds a human leukocyte
antigen (HLA) molecule \citep{wu_ccbhla_2023}. \emph{Metal-ion binding} predicts whether a protein binds
a metal ion \citep{cheng_co-evolution-based_2023}. \emph{Protein--protein interaction} classifies interacting versus
non-interacting protein pairs \citep{bernett_cracking_2024}.

\paragraph{Quantitative activity and variant effects.}
\emph{Enzyme catalytic efficiency} is the benchmark's name for predicting
enzyme turnover ($k_{\mathrm{cat}}$). \emph{$\beta$-lactamase} predicts measured
activity of enzyme variants; \emph{GB1} predicts variant binding-fitness scores;
\emph{fluorescence} predicts fluorescent-protein
brightness~\citep{sarkisyan_local_2016}; and \emph{stability} predicts
resistance to proteolysis~\citep{rocklin_global_2017, xu_peer_2022, dallago_flip_2021, didi_flip2_2026}.

\paragraph{Sequence properties.}
\emph{Solubility} classifies soluble versus insoluble proteins \citep{khurana_deepsol_2018}.
\emph{Signal peptide} predicts the presence of a secretion
signal~\citep{teufel_signalp_2022}, and \emph{neuropeptide precursor}
identifies proteins that yield neuropeptides \citep{Karsenty2014,OferD2014}.
The wall-clock-only \emph{CheZoD} task predicts a disorder score derived from
NMR chemical shifts~\citep{dass_odinpred_2020}.

\paragraph{Residue-level.}
\emph{Secondary structure} assigns each residue one of eight DSSP states;
\emph{disorder} labels each residue as ordered or disordered (NetSurfP-2.0) \citep{klausen_netsurfp-20_2019};
and \emph{signal peptide} labels each residue as inside or outside a signal
peptide (SignalP~6.0) \citep{teufel_signalp_2022}. Each is a linear probe on frozen per-residue
embeddings.

\begin{table}[H]
\centering
\caption{Benchmark sources and headline metrics. Identifiers are Hugging
Face datasets unless a URL is given; \emph{--} marks an absent task; AP is average precision. EC is
multilabel, hence micro-averaged in the controlled suite.}
\label{tab:task-sources}
\small
\renewcommand{\arraystretch}{1.1}
\begin{tabular}{@{}>{\raggedright\arraybackslash}p{0.26\textwidth}>{\raggedright\arraybackslash}p{0.44\textwidth}ll@{}}
\toprule
 & & \multicolumn{2}{c}{Metric} \\
\cmidrule(l){3-4}
Task & Dataset / source identifier & Controlled & Wall-clock \\
\midrule
Remote Homology (Fold) & \nolinkurl{biomap-research/fold_prediction} & Accuracy & F1-Macro \\
SCOPe-40 Retrieval & \nolinkurl{tattabio/scope40_test} & Recall@1 & Recall@1 \\
Subcellular Localization & \nolinkurl{proteinea/deeploc} & F1-Macro & AUC \\
Molecular Function (GO) & \nolinkurl{AI4Protein/GO_MF} & F1-Macro & -- \\
EC Classification & \nolinkurl{AI4Protein/EC} & F1-Micro & F1-Macro \\
Enzyme Catalytic Efficiency & \nolinkurl{biomap-research/enzyme_catalytic_efficiency} & Spearman & Spearman \\
$\beta$-lactamase-PEER & \nolinkurl{SaProtHub/Dataset-Beta_Lactamase-PEER} & Spearman & Spearman \\
Variant Effect (GB1) & \nolinkurl{biomap-research/fitness_prediction} & Spearman & Spearman \\
Fluorescence (TAPE) & \nolinkurl{cradle-bio/tape-fluorescence} & Spearman & Spearman \\
Stability & \nolinkurl{biomap-research/stability_prediction} & Spearman & Spearman \\
CheZoD Disorder & \nolinkurl{PeptoneLtd/contrastive-finetuning-plms} & -- & Spearman \\
Peptide-HLA Binding & \nolinkurl{biomap-research/peptide_HLA_MHC_affinity} & AP & AUC \\
Metal Ion Binding & \nolinkurl{biomap-research/metal_ion_binding} & AP & AUC \\
Protein--protein interaction (PPI) & \nolinkurl{Synthyra/bernett_gold_ppi} & AP & AUC \\
Solubility (DeepSol) & \nolinkurl{proteinea/solubility} & AP & AUC \\
Signal Peptide Prediction & ProteinBERT SignalP\_Binary -- \nolinkurl{https://github.com/nadavbra/protein_bert} & AP & AUC \\
Neuropeptide Precursor & ProteinBERT -- \nolinkurl{https://github.com/nadavbra/protein_bert} & AP & AUC \\
\midrule
\multicolumn{4}{@{}l}{\emph{Residue-level (both studies)}} \\
Secondary Structure (8-state) & \nolinkurl{GleghornLab/SS8} & F1-Macro & F1-Macro \\
Disorder & \nolinkurl{agemagician/NetSurfP-SS3} & MCC & MCC \\
Signal Peptide (per residue) & \nolinkurl{SaProtHub/Dataset-Signal-Peptides} & F1-Macro & F1-Macro \\
\bottomrule
\end{tabular}
\end{table}
\subsection{Next-token prediction on ProGen3-112m}
\label{app:ntp}

\paragraph{Model and objective.}
ProGen3-112m \citep{bhatnagar_scaling_2025} is a causal protein language model with sparse
mixture-of-experts feed-forward blocks (10 layers, hidden size 384, 8 experts,
top-2 routing). We use a PyTorch implementation with converted pretrained
weights. Half of the training sequences are reversed to train both generation
directions. Both arms minimize next-token cross-entropy plus the model's router
load-balancing loss. NTP+JEPA replaces the masked-position latent loss of
Eq.~\ref{eq:loss} with a future-position one: the predictor maps the student's
final hidden state at position $t$ to the EMA teacher's layer-5 (of 10) state at
$t{+}4$, giving the target access to four residues beyond the student's prefix.
Predictor, loss, EMA schedule,
VISReg, and $\lambda$ follow the controlled study.

\paragraph{Training.}
Training follows the controlled study (Sec.~\ref{sec:setup}): UniRef50, 67{,}000
steps, batch 1024, length 512, and the same learning rates and latent-branch
schedule. Validation NTP loss at step 65{,}000 is
2.504 (NTP-only) versus 2.507 (NTP+JEPA) continued, and 2.523 versus 2.532 from
scratch. NTP+JEPA costs 21\% more time per step, so the compute-matched control
continues each NTP-only model for 14{,}070 steps at its final learning rate;
validation loss changes by at most 0.001 nats.

\paragraph{Evaluation.}
We evaluate 16 tasks, excluding PPI, GO, and EC to limit evaluation cost.
Unlike the ESM2 probes, regularization strength is selected per task on the
validation split (or a 10\% training holdout) from
$C\in\{10^{-6},10^{-4},\dots,10^{6}\}$ before refitting on the training data, so
scores are not comparable with Table~\ref{tab:v3-results}. Embeddings are
mean-pooled final-layer states; mean pooling outscored last-token pooling on
average for every arm evaluated with both. Table~\ref{tab:ntp-pertask} lists
per-task scores.

\begin{table}[H]
\centering
\scriptsize
\setlength{\tabcolsep}{3pt}
\caption{ProGen3-112m per-task scores (headline metric; CM: compute-matched).}
\label{tab:ntp-pertask}
\begin{tabular}{@{}lccccccc@{}}
\toprule
Task & Off-the-shelf & NTP & NTP (CM) & NTP+JEPA & Scratch NTP & Scratch NTP (CM) & Scratch NTP+JEPA \\
\midrule
Remote homology & 0.500 & 0.515 & 0.514 & 0.521 & 0.502 & 0.502 & 0.512 \\
Solubility & 0.675 & 0.681 & 0.680 & 0.692 & 0.698 & 0.700 & 0.712 \\
Signal peptide & 0.960 & 0.973 & 0.973 & 0.980 & 0.983 & 0.982 & 0.988 \\
Neuropeptide precursor & 0.967 & 0.986 & 0.986 & 0.974 & 0.989 & 0.988 & 0.987 \\
$\beta$-lactamase & 0.615 & 0.634 & 0.666 & 0.676 & 0.608 & 0.598 & 0.627 \\
Peptide--HLA & 0.771 & 0.808 & 0.809 & 0.835 & 0.813 & 0.814 & 0.844 \\
Metal-ion binding & 0.794 & 0.781 & 0.749 & 0.772 & 0.769 & 0.768 & 0.750 \\
Subcellular loc. & 0.462 & 0.452 & 0.454 & 0.467 & 0.480 & 0.480 & 0.512 \\
GB1 & 0.804 & 0.825 & 0.825 & 0.820 & 0.814 & 0.814 & 0.813 \\
Fluorescence & 0.578 & 0.575 & 0.576 & 0.576 & 0.591 & 0.586 & 0.577 \\
Stability & 0.607 & 0.608 & 0.600 & 0.634 & 0.604 & 0.594 & 0.401 \\
Enzyme efficiency & 0.518 & 0.524 & 0.525 & 0.518 & 0.485 & 0.487 & 0.492 \\
SCOPe-40 R@1 & 0.296 & 0.304 & 0.301 & 0.309 & 0.257 & 0.259 & 0.293 \\
Secondary structure & 0.228 & 0.231 & 0.231 & 0.238 & 0.244 & 0.243 & 0.245 \\
Disorder & 0.268 & 0.275 & 0.274 & 0.306 & 0.292 & 0.291 & 0.296 \\
Signal-peptide residues & 0.590 & 0.602 & 0.601 & 0.609 & 0.605 & 0.605 & 0.617 \\
\bottomrule
\end{tabular}
\end{table}

\subsection{Reproducibility}
\label{app:repro}

Pretraining uses NVIDIA A100 80GB GPUs---one at 35M and two to four at 150M, with
effective batch size held fixed. Controlled-study settings are in Sec.~\ref{sec:setup};
Table~\ref{tab:wallclock-hparams} lists the wall-clock settings. Both studies
start from the official ESM2 weights; the wall-clock study loads them through
Synthyra's weight-identical FastPLMs packaging (\texttt{Synthyra/ESM2-35M},
\texttt{Synthyra/ESM2-150M}).

\begin{table}[H]
\centering
\small
\caption{Wall-clock optimizer and objective settings. Each within-family
contrast shares batch size, masking rate, and schedule. All backbones use
$\lambda=0.45$ and a constant schedule after 1{,}000 warmup steps.}
\label{tab:wallclock-hparams}
\setlength{\tabcolsep}{2.5pt}
\begin{tabular}{@{}lccccc@{}}
\toprule
Backbone & LR (JEPA / MLM-only) & Eff.\ batch & Mask & $\alpha$ & $\lambda$ ramp \\
\midrule
ESM2-35M (pretrained)   & $3{\times}10^{-4}$ / $5{\times}10^{-4}$ & 128 & 25\,\% & $10^{-3}$ & 1k steps \\
ESM2-150M (pretrained)  & $3{\times}10^{-4}$ / $5{\times}10^{-4}$ & 128 & 25\,\% & $10^{-3}$ & 1k steps \\
ESM2-35M (random init)  & $3{\times}10^{-4}$ / $3{\times}10^{-4}$ & 128 & 25\,\% & $10^{-3}$ & 1k steps \\
AMPLIFY-120M            & $5{\times}10^{-5}$ / $5{\times}10^{-5}$ & 208 & 20\,\% & $1.0$ & 2k steps, 25\,\% dropout \\
ProteinBERT2-35M        & $1{\times}10^{-3}$ / $1{\times}10^{-3}$ & 128 & 20\,\% & $1.0$ & 5k delay, 5k ramp, 25\,\% dropout \\
\bottomrule
\end{tabular}
\end{table}

The all-position control uses MSE, target layer normalization, SIGReg, gradient-blocked
clean-input targets, and no JEPA ramp. The masked-position recipe uses the ramps
in Table~\ref{tab:wallclock-hparams} and replaces all-position MSE with cosine
loss at masked tokens. Probes use three seeds, distinct from pretraining seeds.

\paragraph{Controlled-study implementation deviations.}
In the first seed, predictor heads were initialized before seeding, preventing
exact replay; the from-scratch pair also used \texttt{torch.compile} only for
MLM+JEPA. A cross-seed check found no detectable effect on scores.

\paragraph{Wall-clock pretraining seeds.}
Both arms of the masked-position contrast use independent pretraining seeds
(counts below). These runs provide Table~\ref{tab:masked-pos-headline}'s W/L/T,
Fig.~\ref{fig:wallclock-breadth}'s error bars, and pooled SCOPe-40 Recall@1.
Tests pair shared seeds on four backbones; ProteinBERT2 uses an unpaired test.

\begin{center}
\small
\begin{tabular}{@{}lcc@{}}
\toprule
Backbone & Masked-pos MLM+JEPA & MLM-only \\
\midrule
ESM2-35M (pretrained)      & 8 & 8 \\
ESM2-150M (pretrained)     & 7 & 6 \\
AMPLIFY-120M               & 7 & 7 \\
ESM2-35M (random init)     & 7 & 7 \\
ProteinBERT2-35M           & 5 & 6 \\
\bottomrule
\end{tabular}
\end{center}

\noindent
EC uses three runs per arm and the residue-level tasks five to seven. The all-position
EC scores for ESM2-35M and ESM2-150M in Fig.~\ref{fig:wallclock-breadth}
come from independently trained checkpoints of the same recipe, rather than
the checkpoints used for the other tasks.

\noindent
The all-position, JEPA-only, recipe-sweep, and 25{,}000-step runs use one
pretraining seed each; their statistics are task-level.

\end{document}